\documentclass[pmlr,twocolumn,10pt]{jmlr} 
\usepackage{booktabs}
\usepackage{siunitx}
\usepackage{tcolorbox}
\usepackage{svg}
\usepackage{multirow} 
\usepackage{amsmath}
\usepackage{amssymb} 
\usepackage{slashbox}
\usepackage{longtable}

\theorembodyfont{\upshape}
\theoremheaderfont{\scshape}
\theorempostheader{:}
\theoremsep{\newline}

\jmlrvolume{259}
\jmlryear{2024}
\jmlrworkshop{Machine Learning for Health (ML4H) 2024} 

\title{An Interoperable Machine Learning Pipeline for Pediatric Obesity Risk Estimation}

\author{\Name{Hamed Fayyaz} \Email{fayyaz@udel.edu}\\
\addr University of Delaware, Newark, DE, USA
\AND
\Name{Mehak Gupta} \Email{mehakg@smu.edu}\\
\addr Southern Methodist University, Dallas, TX, USA
\AND
\Name{Alejandra Perez Ramirez} \Email{Alejandra.Perezramirez@nemours.org}\\
\addr Nemours Children's Health, Wilmington, DE, USA
\AND
\Name{Claudine Jurkovitz} \Email{CJurkovitz@christianacare.org}\\
\addr ChristianaCare, Newark, DE, USA
\AND
\Name{H. Timothy Bunnell} \Email{Tim.Bunnell@nemours.org}\\
\addr Nemours Children's Health, Wilmington, DE, USA
\AND
\Name{Thao{-}Ly {T. Phan}} \Email{ThaoLy.Phan@nemours.org}\\
\addr Nemours Children's Health, Wilmington, DE, USA
\AND
\Name{Rahmatollah Beheshti} \Email{rbi@udel.edu}\\
\addr University of Delaware, Newark, DE, USA
 }
\begin{document}

\maketitle

\begin{abstract}
Reliable prediction of pediatric obesity can offer a valuable resource to providers, helping them engage in timely preventive interventions before the disease is established. Many efforts have been made to develop ML-based predictive models of obesity, and some studies have reported high predictive performances. However, no commonly used clinical decision support tool based on existing ML models currently exists. This study presents a novel end-to-end pipeline specifically designed for pediatric obesity prediction, which supports the entire process of data extraction, inference, and communication via an API or a user interface. While focusing only on routinely recorded data in pediatric electronic health records (EHRs), our pipeline uses a diverse expert-curated list of medical concepts to predict the 1-3 years risk of developing obesity. Furthermore, by using the Fast Healthcare Interoperability Resources (FHIR) standard in our design procedure, we specifically target facilitating low-effort integration of our pipeline with different EHR systems. In our experiments, we report the effectiveness of the predictive model as well as its alignment with the feedback from various stakeholders, including ML scientists, providers, health IT personnel, health administration representatives, and patient group representatives. 
\end{abstract}

\paragraph*{Data and Code Availability}
Our code, containing the application model with parameter (weight) values, is publicly available on GitHub at \url{https://github.com/healthylaife/fhir}. A live demo of our proposed application is available at \url{https://fhir-obesity.com}. The tool will also be available on SMART App Gallery at \url{https://apps.smarthealthit.org/}.

\paragraph*{Institutional Review Board (IRB)}
Our research has obtained two separate IRB approvals from a panel at Nemours Children's Health for tool development and human evaluation.

\section{Introduction}
Pediatric obesity is a major public health problem that has been escalating across the globe \citep{pulgaron2013childhood}. Specifically, in the US, the prevalence of obesity in children and adolescents aged 2-19 years in 2017-2020 was 19.7\% and affected about 14.7 million children and their families \citep{akinbami2022national}. This disease is often comorbid with a variety of health issues, such as type 2 diabetes, hypertension, and sleep apnea \citep{pulgaron2013childhood}; and increases the risk of obesity-related complications in adulthood, including heart disease, stroke, and certain types of cancer \citep{reilly2011long}. Beyond the physical health implications, pediatric obesity can lead to psychological problems like low self-esteem, depression, and social isolation \citep{rankin2016psychological}. Addressing this public health problem requires a multidisciplinary approach involving parents, educators, healthcare professionals, and policymakers to create an environment supporting children's healthy lifestyles. Timely identification of patients at an elevated risk of obesity can help engage in preventive interventions before the disease is established. In particular, utilizing AI methods to predict unhealthy weight trajectories can offer a more targeted approach and allow for early interventions and more efficient treatment in tackling this complex disease.

Despite the abundance of obesity predictive models \citep{graversen2015prediction, redsell2016validation, steur2011predicting, weng2013estimating, hammond2019predicting}, existing models are not commonly used in clinical practice. One major limitation of existing work relates to using ``site-specific'' variables, such as ad hoc datasets and questionnaires, in carefully crafted data collection studies \citep{gupta2022obesity}. This kind of data is unavailable to all pediatric care providers.

Furthermore, from an interoperability point of view, bridging the gap between theoretical models and real-world implementation within the healthcare domain has proven to be another major challenge \citep{harris2022clinical}. Smooth translation of technical tools requires integrating the models into the workflow of pediatric providers with minimal modification and adjusting. Collaborative efforts between data scientists, stakeholders, healthcare professionals, and technology experts are also needed to ensure the successful deployment and application of obesity prediction systems in clinical settings.

One way to integrate such ML tools into clinical practice is by using Fast Healthcare Interoperability Resources (FHIR) protocols \citep{ayaz2021fast}, which provide a standardized framework for exchanging healthcare information. Implementing FHIR allows for seamless interoperability between different health information systems, facilitating the smooth data exchange between predictive obesity models and EHR platforms. 



This study aims to present an end-to-end application that can access EHR data through FHIR protocols, extract relevant variables, and predict the risk of developing obesity in the future.
Specifically, we propose an application that processes the EHRs of a patient and predicts obesity using greatly customized machine learning methods. It also addresses the common challenges of working with EHRs, including missingness, high-dimensionality, and temporality. Our pipeline works with only a subset of electronic health records (EHRs) that are routinely recorded in pediatric healthcare systems and avoids using site-specific or uncommon medical concepts.  
Specifically, the contribution of our study can be listed as follows:

\begin{itemize}
\item We propose a comprehensive obesity prediction pipeline that can be integrated into any EHR platform using FHIR protocols. We also present a dedicated API for retrieving health records and returning obesity predictions.
\item We use our deep learning model customized for learning longitudinal BMI trajectories from commonly available EHR data elements to predict obesity for the next three years. 
\item We present a dedicated user interface designed following extensive collaboration with different stakeholders, including the pediatric providers (i.e., the end users). 
\end{itemize}


\section{Background}
Two research areas are related to our work, namely, obesity prediction using machine learning and clinical decision support systems. In this section, we review existing work in these areas.

\paragraph{Obesity Prediction}
Machine learning models have shown great potential in predicting future health outcomes. The availability of large-scale medical data has made clinical predictive models prevalent and widely utilized in the medical domain \citep{alonzo2009clinical}. 


In the field of obesity prediction, a growing subset of studies have used advanced machine learning techniques. Most of the existing studies rely on traditional machine learning methods, such as the studies that used regression \citep{graversen2015prediction, redsell2016validation, steur2011predicting, weng2013estimating}, random forest \citep{hammond2019predicting} or sequence mining \citep{campbell2020identification}. Capturing temporal and non-linear patterns in the complex EHR data is, however, not straightforward through these techniques. Some other studies have used advanced deep learning models for obesity prediction \citep{ziauddeen2018predicting, colmenarejo2020machine, thamrin2021predicting, zhou2022applications, dutta2024obesity,Mottalib, fayyaz22a,gupta2024associations}. These studies are extensively reviewed by \citep{ferreras2023systematic, yi2024review}. 

These studies often use pre-selected and site-specific variables like maternal and self-reported data to predict obesity at future time points.


\paragraph{FHIR-based Clinical Decision Support Systems}
Clinical decision support (CDS) systems play a prominent role in enhancing health delivery and practice by providing critical information to healthcare professionals in various stages of patient care. These systems assist clinicians in making decisions by leveraging patient data, medical knowledge, and best practices. CDS systems have numerous applications, especially when integrated with the providers' routine workflow. Some notable examples include estimating life expectancy in cancer patients \citep{Koutkias2019Contributions}, treatment of depression \citep{Borbolla2021Clinical},  and managing chronic diseases \citep{souza2020clinical}.

Some recent generations of CDS systems have adopted FHIR protocols to integrate predictive machine learning tools into EHR systems \citep{balch2023machine, sun2022machine}. By integrating with EHRs, CDS systems can utilize comprehensive health data of patients and offer timely information, such as best practice alerts (BPAs). Furthermore, the interoperability between CDS systems and EHRs can be facilitated by standards like FHIR, which brings about seamless communication and data exchange. Traditional CDS systems and EHR systems with ad-hoc communication protocols have faced persistent challenges in long-term maintenance or external adoptions \citep{dullabh2022challenges}. FHIR-based CDS systems can potentially be integrated into (almost) any EHR system with nominal setup requirements, as FHIR enables the structured sharing of clinical data, allowing CDS systems to access and analyze patient information efficiently \citep{balch2023machine}.

While we are not aware of any similar study for obesity prevention or management, a range of studies have also explored the use of machine learning methods in a FHIR-based setting for specific medical applications such as chronic obstructive pulmonary disease \citep{curran2020integrated}, diabetes \citep{10.1093/jamiaopen/ooz056, tarumi2021leveraging}, asthma \citep{thayer2021human}, transfusion allergic reactions \citep{whitaker2022detection}. Furthermore, methods for transforming existing CDS systems into FHIR structures have been discussed in prior work \citep{KOVALENKO2019Transformation, gruendner2019ketos}.




\begin{figure}
    \centering
    \includegraphics[width=0.95\linewidth]{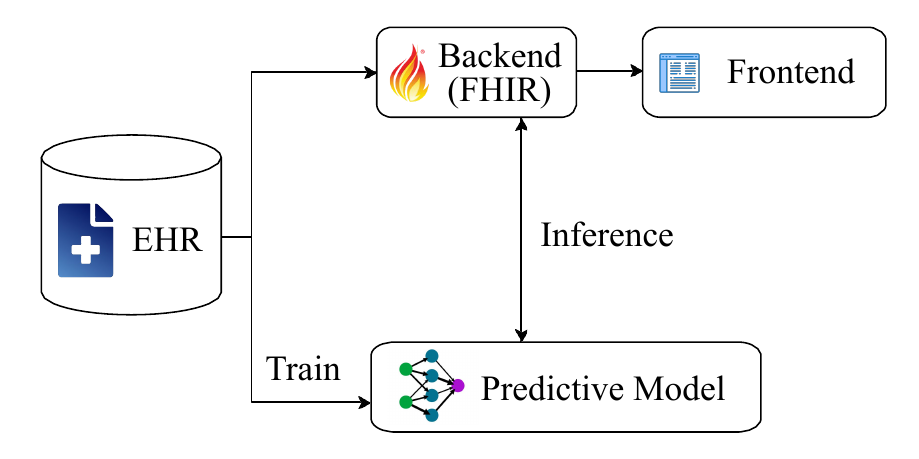}
    \caption{Overview of the proposed system. It consists of three main components. (1) A predictive model that has been trained using EHR data to predict the risk of obesity in the next three years. (2) A backend that uses the model to predict the risk of obesity for new patients. (3) A frontend (user interface) for expert interaction with the system.}
    \label{fig:overview}
\end{figure}

\section{Method}
\label{sec:method}
Our obesity prediction pipeline includes three main components. (1) A backend that communicates with the EHR database in Observational Medical Outcomes Partnership (OMOP) common data model (CDM) format using the FHIR protocol and OMOP-on-FHIR to obtain patients' medical records. OMOP-on-FHIR is a FHIR server implementation built on top of the OMOP CDM. It is designed to provide a FHIR clinical API, which can read and write data from an OMOP CDM database. It also includes a preprocessing pipeline to transform the data representation in a way that is ready for the obesity predictive model. The backend can be accessed through the user interface or the dedicated API.  (2) A trained machine learning model that predicts obesity for each patient for the next three years. (3) A user interface based on SMART (Substitutable Medical Apps and Reusable Technologies)-on-FHIR that shows the summary of patient health records contributing to the prediction of obesity as well as the outcome of the obesity predictor model. SMART-on-FHIR is a set of open standards and specifications designed to enable the integration of health information technology systems in a secure and interoperable manner. SMART-on-FHIR combines the SMART platform with the FHIR standard to create a framework for building healthcare applications. The overall architecture of the proposed system is shown in Figure \ref{fig:overview}. We discuss the details of these components below.

\subsection{Backend}
The backend acts as a coordinator between the user interface, data sources, and machine learning model. It comprises three components: (1) a FHIR client, (2) an API, and (3) a preprocessing pipeline. The FHIR client (1) communicates with the EHR databases using FHIR protocols to receive raw data. The API  (2) can be used by the user interface or independently to send the patient data resources and receive the predictions from the obesity predictor model. The preprocessing pipeline (3) transforms the data representation to match the expected format by the machine learning model.

\begin{figure}[h]

  \centering
  \includegraphics[width=0.99\linewidth]{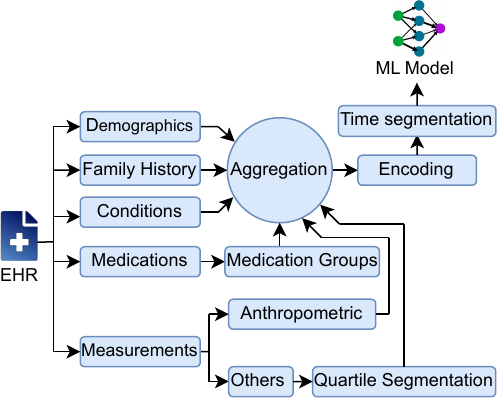}
  \caption{The data flow in the pipeline. 
  }
  \label{fig:dataflow}
\end{figure}

\subsubsection{FHIR Client}
We used the SMART-on-FHIR framework to build our FHIR client, which is a platform-agnostic platform \citep{mandel2016smart}. The FHIR client receives the concepts (i.e., demographics, measurements, medications, conditions, family history) for each patient from the EHR database and sends them to the preprocessing pipeline for further steps.

Using FHIR standardization promotes data consistency and compatibility, enabling efficient stakeholder communication and collaboration \citep{vorisek2022fast}. Leveraging FHIR protocols can streamline integrating technical models into practical healthcare settings, thus enhancing the overall usability and effectiveness of obesity prediction systems in pediatric clinical environments. 

\subsubsection{API}
We used Flask \citep{grinberg2018flask} to develop our API. Flask is a lightweight Python web framework designed for building web services and applications that provide essential tools for rapid development. Its modular structure and adherence to the Web Server Gateway Interface standard make integrating with various web servers easy.

Our API can be utilized through the user interface or independently for other downstream applications (i.e., using obesity risk prediction as an input for other CDS tools). The API can access the FHIR resources directly or receive them in JSON format. In the former mode, it receives a patient's identification number and uses the FHIR client embedded in the application to retrieve the patient's information. In the latter mode, the API directly receives the FHIR resources in JSON format. The data is then passed to the preprocessing pipeline for further steps.

\subsubsection{Data Preprocessing Pipeline}
As mentioned earlier, we adopted a variant of the OMOP CDM as implemented for the PEDSnet clinical research network (NEED REFERENCE), which is a data standard designed to standardize the structure and content of observational data and enable efficient analysis. OMOP transforms data from various sources, such as EHRs and administrative claims, into a common format with standardized vocabularies and coding schemes. The CDM includes a person-centric relational database schema with a group of tables categorized into domains. It captures a wide range of healthcare information, including patient encounters, diagnoses, drugs, and procedures. As an open-community standard, the adoption of the OMOP CDM can facilitate reproducible research and collaboration across institutions.


\paragraph{Specific procedure for obesity prediction} For the initial training of our models, we use data collected from the Nemours Children's Health system, which is a large network of pediatric healthcare facilities in the eastern US. This dataset is part of the PEDSnet network with EHR data from 11 large pediatric health systems \citep{forrest2014pedsnet}. Our data is chosen from over two million patients from the Nemours portion of PEDSnet EHR system. For training, we included patients in the cohort who (1) have at least 5 years of medical data, (2) at least one BMI was recorded in their medical history data, (3) no evidence of type 1 diabetes, (4) no evidence of cancer, sickle cell disease, developmental delay, or other complicated medical conditions. The latter two exclusion criteria relate to especially complex (requiring intensive treatment, multiple hospitalizations, etc.) and our data center’s standard protocol is to exclude these populations from most data extractions unless they are the population of interest for a particular study.

\begin{figure}
    \centering
    \includegraphics[width=1\linewidth]{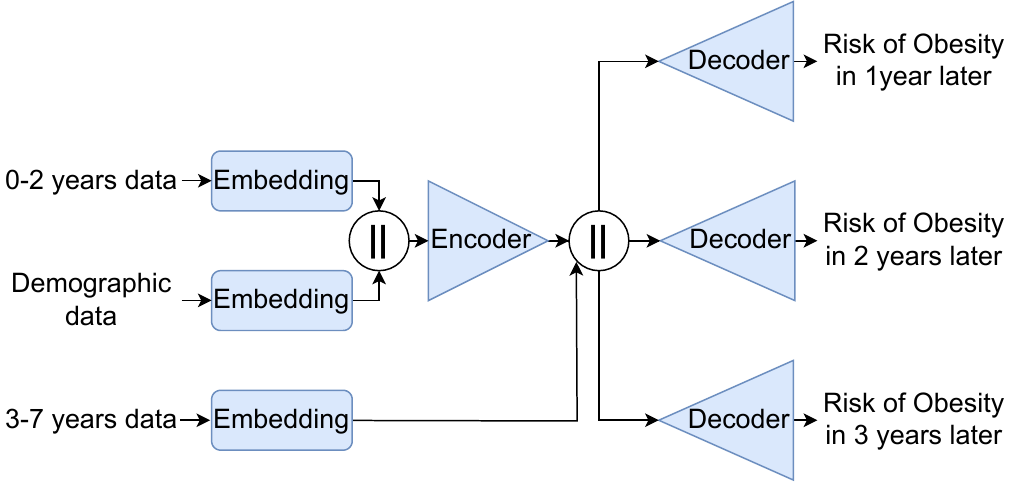}
    \caption{An overview of the machine learning model, using a sample scenario of using 0-2 yr data to predict obesity at 3-7 yr. The encoder (Enc) is a two-layer LSTM network, and the decoder (Dec) consists of two fully connected layers. $||$ shows concatenation.}
    \label{fig:model}
\end{figure}

The BMI of the majority of patients was in the normal range. Therefore, we undersampled the normal weight population to balance the dataset. The dataset was anonymized. All dates were randomly skewed per patient by $\pm 180$ days. The data access and processing steps were approved by the Nemours Children's Health institutional review board (IRB). 

The dataset included demographic data as well as inpatient and outpatient visit information, conditions, procedures, medications, and measurements that are related to each visit. Standard SNOMED-CT, RxNorm, CPT, and LOINC medical coding were used to encode health information in the original dataset. Additional information about the dataset is shown in Appendix (\ref{sec:cohort}).

\paragraph{Data Representation}
EHR data contains patient records as a sequence of visits over time, with each visit containing a variety of medical codes. To represent this data, we use an incremental approach with multiple levels of representation. First, we represent medical codes using code-level representation, including conditions, medications, procedures, and measurements. Then, we combine the code-level representations into visit-level representations, representing all the medical codes for each patient visit. Finally, we use visit-level representations to create a patient-level representation, combining all visits to represent the full sequence for each patient. 

In addition to temporal data for the visits, EHRs also contain static demographic data, which does not change with every visit. We show the demographic variables such as race, ethnicity, gender, insurance type, and address as categorical variables. 
We preprocess and aggregate all these data types for further processing.


Based on input from a group of pediatric obesity experts in our team, we selected a subset of clinically relevant features that are generally recorded in pediatric (well-)visits. This selection process yields 71 clinical diagnoses, 67 family history diagnoses, 84 medication groups, and 51 measurements (the exact list is available in  \citep{GUPTA2024100128}). Taking the above steps transforms the initial sparse representation to a representation with appropriate density for further processing.

We quantize the measurement variables that are initially present as continuous variables. All condition and procedure variables are converted to binary values, with 1 indicating the variable is present for the visit and 0 indicating the variable is not recorded. Then, we aggregate and unify all features with different types in one table, including age, type, description, and value.

Additionally, we encode the data by assigning a unique identifier to each feature. We also divide the patient's timeline into time bins and aggregate features over each time bin. We divide the timeline into equally sized monthly bins (for a  0--2 yr period with more visits) and bimonthly bins (for later ages). 
An overview of the data flow in the preprocessing pipeline is shown in Figure \ref{fig:dataflow}.


\begin{figure*}[hbtp]
    \centering
    \includegraphics[width=0.75\linewidth]{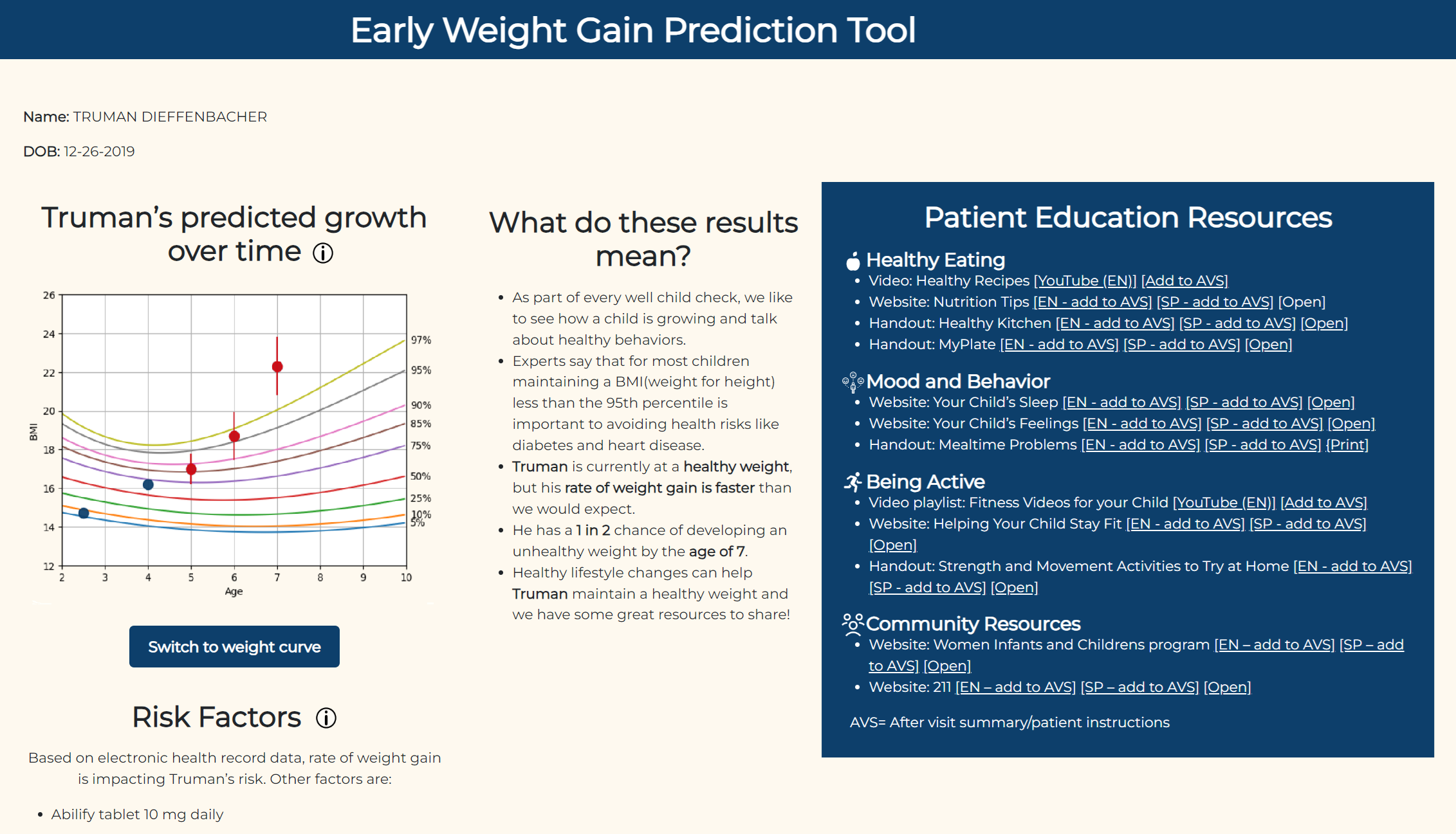}
    \caption{The user interface of our pipeline for a hypothetical (synthetic) patient, showing the predictions of the model and the resources the providers and families can use. }
    \label{fig:application}
\end{figure*}

\subsection{Machine learning model}

We extended the encoder-decoder architecture presented in our previous work \citep{gupta2022flexible}. We have extensively evaluated our ML model through discriminative performance checking (including comparison to other baselines), temporal (different times for training and testing), and geographic (using data from other sites) validations. We have calibrated our model and checked for its fairness performance across various protected groups (determined by gender, race, and insurance type)\citep{GUPTA2024100128,gupta2022obesity,Gupta22concurent}.

The extended version includes an encoder consisting of embedding layers that map diagnosis, medication, procedure, and measurement data into a latent space with 256 dimensions, followed by two layers of LSTM cells with 512-dimension neurons. 

Consider a scenario where a patient has the data from birth to age 7 yr. As shown in Figure \ref{fig:model}, the encoder takes EHR data for 0-2 yrs as input and outputs a representation (embedding) vector. All the features in 0-2 yrs of EHR data are arranged chronologically, where the order of events occurring at the same timestamps is random. The output vector capturing visit-level (temporal) patterns is concatenated with the demographic data representation, where the demographic data is embedded into the latent space using a separate embedding layer. 

This concatenated vector is given to the decoder as input. The decoder can learn from medical data of different lengths. In the example scenario, the decoder concatenates this vector with the 3-7 yrs EHR data. 
%
The decoder contains three separate feed-forward networks with two fully connected layers (512-dimension with Leaky Relu activation of 0.1 slope, 256-dimension, and 0.2 dropout rate) for every future age point for the next three years. A softmax layer at the end of two fully connected layers generates the final output. Each of these three separate feed-forward networks is used to simultaneously provide the risk of obesity for the next 1, 2, and 3 years. Obesity is defined based on the CDC's standard thresholds of being above the 95\textsuperscript{th} percentile for corresponding age and sex.

Feature importance values for all features are obtained by calculating the attention scores. Since we use different decoders for each year's prediction, the scores obtained can be different for features from each task. For instance, in the presented user interface in Figure \ref{fig:application}, we rank top contributors from age 0 to 2 years for all prediction tasks. 


\subsection{Frontend (User Interface)}
The user interface (shown in Figure \ref{fig:application}) is designed for the end-users interacting with patients and their families, in particular pediatric primary care providers. The user interface consists of several components that provide essential information about the patient's medical history and future weight prediction based on an embedded machine learning model. 

The final configuration of the user interface was set up following an extensive user study, which we discuss the details of it later in Section \ref{sec:exp}. The user interface has been designed in close partnership with our support group, which includes diverse stakeholders, primary care providers, and human computer interaction experts. While we could have shown a wealth of information through our AI, we have especially aimed for simplicity and briefness, considering the very busy workflow of pediatric providers who already deal with busy panels and patient charts during the visits. 

The final components of the user interface include the following five. (1) Patient's information (name and date of birth). We do not include race, ethnicity, and address. (2) A BMI trajectory displaying the child's BMI history and predictions for the next three years. The BMI values are superposed over the standard CDC growth charts, creating a familiar interface to what providers generally see in patient charts.  We have also designated an option to switch between the weight and BMI trajectories. (3) A summary of the top risk factors as determined by the predictive ML model (as explained in \cite{gupta2022flexible}, including top medications, measurements, family history, and conditions. (4) A descriptive guide is presented to help the users understand and interpret the outputs of the model, assist in interpreting the predictions, and offer advice on how to navigate each case. (5) A comprehensive panel listing various educational resources for families and their children is also included. These resources cover nutrition, health behavior, and community resources that can help prevent obesity and promote a healthier lifestyle.


\section{Experiments}
\label{sec:exp}
We conducted two types of experiments to evaluate the effectiveness of our developed pipeline, targeting the performance and utility of the whole pipeline. In particular, we target a series of technical experiments to asses the overall prediction (discrimination) performance of our ML model. We also conduct experiments that target the human factors involved in designing our pipeline, primarily focusing on the usability of the user interface within the pediatric clinical workflow.

\subsection{Machine Learning Model Evaluation}
We trained our deep learning model using 36,191 eligible children to predict the risk of obesity as well as BMI in the next three years using 6 different lengths of observation windows from ages 0-2 to 0-7 years. The observed Area Under the Receiver Operating Characteristic Curve (AUROC) for all prediction ages is presented in Table \ref{tab:result}.  

At a high level, there are trade-offs between the length of prediction windows and the accuracy of the models. Longer windows are preferred but can reduce accuracy. Shorter windows risk not giving enough time for an effective intervention. The current granularity of predictions (next 1-3 years) was chosen based on input from the pediatric obesity experts in our team and considering the common frequency of pediatric visits after age 2 (annual).

Specifically, during the first two years, children typically have well-child visits at 1, 2, 4, 6, 9, 12, 18, and 24 months. Following this period, annual visits are recommended. Based on this schedule, we utilized 3-month granularity in the first year and 6-month granularity in the second year. After two years, we adopted a yearly data granularity, which still captures sufficient variability in BMI to provide meaningful signals for the model's learning process. We will add this information to the manuscript.

We report additional results related to performance across temporal (different years), geographical (two different regional sites), and various subgroups (as determined by gender, race/ethnicity, and insurance type), as well as net benefit analysis, in the Appendix \ref{sec:extended_result}.

\begin{table}[ht]
\centering
\resizebox{\columnwidth}{!}{
\begin{tabular}{l|ccc}
\toprule
  & \multicolumn{3}{c}{Predict Next}\\ \cline{2-4} 
Input Yrs  & 1 yr&  2 yr & 3 yrs\\
\cline{2-4}
& \multicolumn{3}{c}{BMI over 95th Percentile (Classification)}    \\
\midrule
0-2             & 0.87 (.02)     & 0.84 (.02)     & 0.81 (.02)     \\
0-3             & 0.90 (.02)     & 0.88 (.02)     & 0.84 (.02)     \\
0-4             & 0.91 (.02)     & 0.88 (.02)     & 0.86 (.02)     \\
0-5             & 0.92 (.02)     & 0.90 (.02)     & 0.87 (.02)     \\
0-6             & 0.92 (.02)     & 0.90 (.02)     & 0.88 (.02)     \\
0-7             & 0.93 (.01)     & 0.91 (.03)     & 0.87 (.02)    \\
\midrule
& \multicolumn{3}{c}{Exact BMI Value (Regression)}    \\
\midrule
0-2             & 0.91 (1.72)       & 0.95 (1.77)     & 1.07 (2.30) \\
0-3             & 0.74 (1.65)       & 0.92 (2.00)     & 1.07 (2.80) \\
0-4             & 0.86 (1.72)       & 1.09 (2.56)     & 1.35 (3.18) \\
0-5             & 0.86 (1.81)       & 1.13 (2.48)     & 1.5 (3.54)  \\
0-6             & 1.02 (2.22)       & 1.45 (2.86)     & 1.76 (3.54) \\
0-7             & 1.27 (2.33)       & 1.55 (2.99)     & 1.92 (3.60) \\
\bottomrule
\end{tabular}
}
    \caption{Predictive performance of the ML model. The top part shows the classification performance in predicting whether a child would be above the 95th percentile or not. Mean AUROCs (95\% CI) are shown. The bottom part shows regression performance in predicting future BMI values. Mean absolute errors (95\% CI) are shown. The uncertainty range calculations have been done using conformal prediction.
    }
    \label{tab:result}
\end{table}

\subsection{User Study}
We recruited a group of 12 primary care providers (PCPs) from our pediatric healthcare system. We have obtained extensive qualitative feedback about the usability, acceptability, and utility of our developed tool to predict early childhood obesity risk. 

In our iterative study, we have assessed the acceptability and feasibility of the refined CDS tool among the PCPs using the Technology Adoption Model (TAM) \citep{holden2010technology} questionnaire. This study was approved in a separate IRB process (besides the tool development) by a local board at Nemours Children's Health.

Enrolled providers were asked to provide feedback on the CDS tool prototype. Participants met with a trained research coordinator via secure video conference. Providers were shown the prototype and asked questions about the usability, acceptability, and utility of each component of the tool and the overall tool itself using a semi-structured interview guide. After each session, provider feedback was summarized utilizing a modified cognitive interview analysis protocol \citep{knafl2007analysis}.

In addition to reaching saturation of responses during usability testing, participants were asked to provide an evaluation of the final CDS prototype. They were sent a screenshot of the final CDS prototype (not linked to or presenting real data) and the TAM questionnaire to complete.  TAM is a validated questionnaire that has been used in the evaluation of healthcare information technology. It consists of 6 items assessing both the usability and utility of a tool, with responses to each item being on a 5-point Likert scale (Strongly Disagree -- Strongly Agree). All questionnaire responses were anonymous. 

Table \ref{tab:survey_data} shows the overall results we have obtained through our user study. The average TAM score was 4.45 (SD 0.54), with 7 out of 10 users scoring an average TAM score greater than 4 (the other 3 scored between 3.7 and 3.8). While the completion of demographic information was optional per our IRB, 7 participants provided it: 6 were female, 1 male; 5 were White, 1 Hispanic, and 1 Asian—all of whom were physicians. The average number of years in practice among them was 11, and one had specific training in nutrition or obesity. The developed questionnaire for our study is available in Appendix \ref{q}. The interview guide for user testing of the obesity prediction tool is also presented in Appendix \ref{apx:int}.

\begin{table}[h]
\centering
\begin{tabular}{ll}
\toprule
Number of participants               & 12              \\ \midrule
\multirow{2}{*}{Gender Distribution}    & Female: 6    \\
                                        & Male: 1      \\ \midrule
\multirow{3}{*}{Ethnicity Distribution} & White: 5     \\
                                        & Hispanic: 1  \\
                                        & Asian: 1     \\ \midrule
Years in Practice (mean)                  & 11             \\ \midrule
TAM Score (mean)                      & 4.45 (SD 0.54) \\

\bottomrule
\end{tabular}
\caption{Summary of Survey Responses. For demographic information, only the reported results are shown.}
\label{tab:survey_data}
\end{table}

The majority of participants (11/12 $\approx$ 92\%) found the tool to be useful and acceptable. Qualitative feedback informed iterative adaptations to the user interface.  In particular, adaptations were made to include easily understood growth curves, interpretation of the results using family-friendly language, and formatting so all important elements could be accessible on the same screen. After iterative evaluation and updating of the tool, saturation (majority of participants accepting the version) was reached on the final user interface shown in Figure \ref{fig:application}.

\section{Discussion}
A fairly large group of machine learning models have been presented for predicting future patterns of obesity across various settings (e.g., pediatric vs. adult, prevention vs. treatment, and using self-report vs. EHR data). As in many other predictive health AI tools, most predictive models for obesity are not deployed in clinical environments \citep{habehh2021machine}. One key barrier to the wide adoption of such tools is using site-specific designs, such as specific input variables and data formats \citep{adler2022meeting}. In this work, we designed and implemented a pipeline for integrating our obesity prediction model in the clinical setting while making interoperability across various clinical settings a top priority.


To this end, we developed our end-to-end pipeline, which consists of data preprocessing, predictive modeling, and user interface, following FHIR protocols. 
Using FHIR protocols for data exchange can greatly reduce the barriers to the wide adoption of our pipeline across different healthcare systems. 
Our pipeline can meaningfully help healthcare professionals detect patients with a high risk of developing obesity in advance, therefore helping the patients and their families make more informed decisions to reduce the risks of obesity. Prior research has shown the potential of effective communication with children in reducing the risks of obesity \citep{cheng2022communicating}. 

In our experiments, we have demonstrated the effectiveness of our pipeline in estimating future risks of developing obesity. In particular, our model consistently achieved an area under the curve (AUROC) of above 0.8 (with most cases around 0.9) for predicting obesity within the next 3 years for children 2 to 7. The validation results show the robustness of the model. 

We have also run an extensive user study by working closely with various stakeholders, especially pediatric PCPs.  We conducted iterative Think Aloud testing until no further substantive recommendations were provided. Changes were made with each round of testing to address substantive recommendations. The themes that emerged during testing included the need to use clinically relevant visuals and family-friendly language to help clinicians explain the results to families, provide healthy lifestyle resources for clinicians to give to families after discussing the results, and simplify the UI to increase efficiency in using the tool. For example, clinicians wanted everything on one page, so we created all the drop-downs and information buttons, and we used growth curves, which is what clinicians are used to explaining to families. 

A natural next step of our project would be a prospective evaluation of the pipeline either in silent (not used for CDS) or pilot (deployed in small and controlled scale) modes. Our team is actively working toward these next steps and we hope other teams will also adopt our public tool and join us in our fight against pediatric obesity. 

\textbf{Limitations:} Our study remains limited in several ways. The model's performance may be influenced by the quality and completeness of the data. Variations in data quality across EHR systems could potentially impact the accuracy and reliability of predictions, leading to challenges in generalizability. By adhering to FHIR protocols and leveraging the OMOP common data model, our system aims to bridge the differences and facilitate seamless integration regardless of the underlying EHR infrastructure. We have tested our pipeline on Epic through the ``EPIC on FHIR'' interface and confirmed its integration. 

Moreover, inherent biases in the training data (such as social and measurement biases) or feature-matching processes could affect its performance. Additionally, while the model's FHIR compatibility facilitates user accessibility and integration with EHR systems, the long-term usability and acceptance of our application among healthcare providers and patients would require further evaluation and refinement to ensure effective adoption in clinical practice and maintain positive net benefit.   The issue of a performance drop and effective implementation of ``continual learning'' approaches is an important consideration that applies to all AI-based CDS tools. 

Finally,  long-term prediction of pediatric obesity may pose additional challenges due to potential shifts in the underlying data patterns over time. Factors such as evolving healthcare practices, changes in patient demographics, and advancements in medical technologies could introduce variability in the predictive performance of the model over extended periods. Consequently, continuous monitoring and recalibration of the models may be necessary to ensure the reliability and accuracy of the models in forecasting obesity risk in children over an extended period.

\acks{Our study was supported by the NIH award U54-GM104941 and computing credits from Amazon Web Services
 (AWS). We would like to acknowledge  significant support from Haritima Manchanda (U. Delaware), 
Matthew Mauriello (U. Delaware),
Levon H Utidjian (CHOP), Daniel Eckrich  (Nemours), and all PCPs who helped us in developing our pipeline.}

\bibliography{main}
\clearpage
\appendix

\section{Dataset Demographics}
\label{sec:cohort}
Table \ref{tab:cohort} shows the distribution of some of the demographic variables in the utilized dataset.
\begin{table}[!h]
    \label{tab:cohort}
    \caption{Characteristics of the cohort used in this study}
    \centering
    \begin{tabular}{lll}
        \toprule
         \multicolumn{2}{l}{Name}                            & Value \\
         \midrule
         \multicolumn{2}{l}{Total \# of patients}            & 68,003 \\
         \multicolumn{2}{l}{Total \# of visits}              & 3,334,047 \\
         \multicolumn{2}{l}{Avg. \# of visits per patient}   & 51\\
         \midrule
         \multicolumn{2}{l}{Age}                             & [0-20], Mean=5\\
         \midrule
         \multirow{2}{*}{Gender}&  Male                      & 31,003 (45\%) \\
                                &  Female                    & 37,000 (54\%) \\
         \midrule
         \multirow{2}{*}{Race}  &  White                     & 33,244\\
                                &  Black                     & 25,329\\
                                &  Non-Hispanic              & 58,894\\
                                &  Other                     & 17,834\\   
         \bottomrule
    \end{tabular}
\end{table}


\section{Extended Results}
\begin{figure}
    \centering
    \includegraphics[width=1\linewidth]{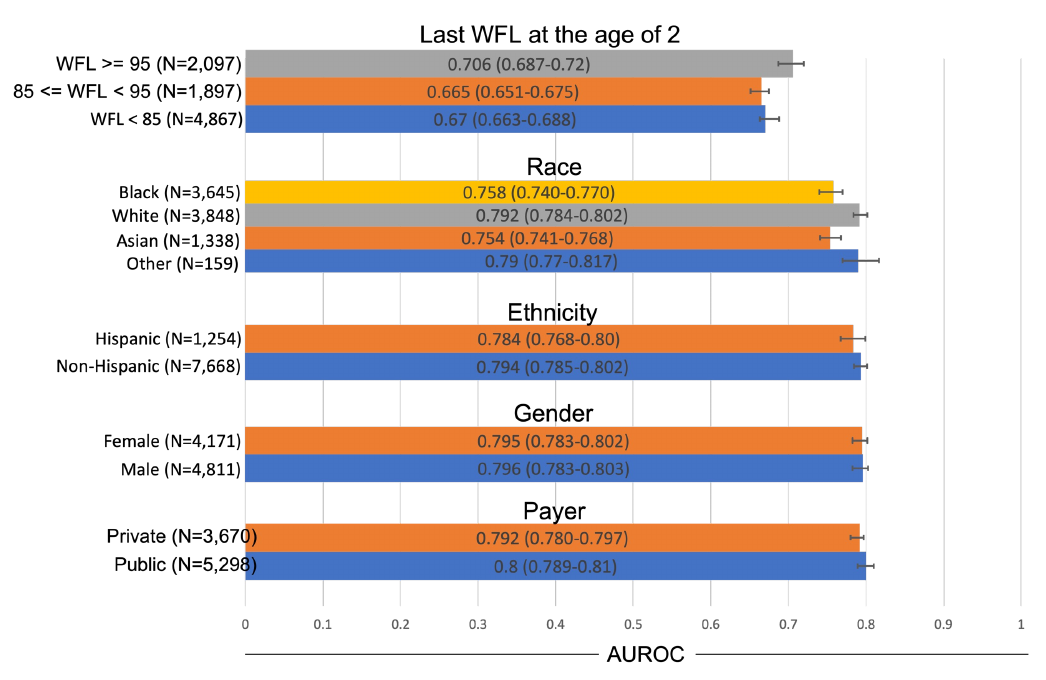}
    \caption{Evaluating the predictive model's robustness by comparing AUROCs (in the test dataset) across five groups (13 subgroups): last WFL before age 2 (3 categories), race (Asian, Black, White, Other), ethnicity (Hispanic, Non-Hispanic),  sex (female, male),  and payer (private, public).}
    \label{fig:subgroup}
\end{figure}

Additional results related to performance across temporal (different years), geographical (two different regional sites), and various subgroups (as determined by gender, race/ethnicity, and insurance type), as well as net benefit analysis, are shown in Table \ref{tab:extended_Result} and Figure \ref{fig:subgroup}.

We validated our model temporally and geographically and studied the model's performance across subpopulations (which can also capture the model's fairness across these groups) using separate test datasets. When using the entire cohort, the data was split with an 80:20 train and test regime, with 5\% of the training data as a validation set to fix the best model. Model performance was reported exclusively on the test dataset. The confidence intervals (CIs) for the classification task are calculated using 100 bootstrapped replicates. For the regression task, we used conformal prediction. 

\label{sec:extended_result}
\begin{table*}[ht]
\centering
\begin{tabular}{lllll}
\toprule
\textbf{Input Age Range} & & \multicolumn{3}{c}{\textbf{PredictionAge}} \\
\midrule
0-2     &                           & 3               & 4               & 5               \\
\midrule
        & AUROC(95\%CI)             & 0.87(0.85-0.88) & 0.84(0.81-0.86) & 0.81(0.79-0.83) \\
        & Temporal validation       & 0.85(0.84-0.86) & 0.83(0.82-0.84) & 0.81(0.80-0.82) \\
        & Geographic validation     & 0.82(0.80-0.83) & 0.80(0.79-0.82) & 0.77(0.76-0.79) \\
        & Net Benefit(20/40/60)     & 0.14/0.09/0.05  & 0.12/0.07/0.04  & 0.12/0.08/0.04  \\
\midrule
0-3     &                           & 4               & 5               & 6               \\
\midrule
        & AUROC(95\%CI)             & 0.90(0.87-0.92) & 0.88(0.86-0.90) & 0.84(0.82-0.86) \\
        & Temporal validation       & 0.89(0.87-0.91) & 0.86(0.84-0.88) & 0.83(0.81-0.85) \\
        & Geographic validation     & 0.85(0.84-0.87) & 0.83(0.81-0.84) & 0.79(0.77-0.80) \\
        & Net Benefit(20/40/60)     & 0.17/0.12/0.08  & 0.14/0.09/0.05  & 0.12/0.07/0.03  \\
\midrule
0-4     &                           & 5               & 6               & 7               \\
\midrule
        & AUROC(95\%CI)             & 0.91(0.89-0.92) & 0.88(0.86-0.90) & 0.86(0.84-0.87) \\
        & Temporal validation       & 0.90(0.88-0.92) & 0.86(0.85-0.88) & 0.84(0.82-0.86) \\
        & Geographic validation     & 0.87(0.85-0.89) & 0.84(0.82-0.85) & 0.82(0.80-0.84) \\
        & Net Benefit(20/40/60)     & 0.16/0.11/0.08  & 0.12/0.09/0.05  & 0.12/0.08/0.05  \\
\midrule
0-5     &                           & 6               & 7               & 8               \\
\midrule
        & AUROC(95\%CI)             & 0.92(0.90-0.93) & 0.90(0.88-0.92) & 0.87(0.85-0.88) \\
        & Temporal validation       & 0.91(0.89-0.92) & 0.88(0.87-0.89) & 0.85(0.84-0.86) \\
        & Geographic validation     & 0.89(0.87-0.91) & 0.85(0.83-0.86) & 0.83(0.82-0.84) \\
        & Net Benefit(20/40/60)     & 0.17/0.12/0.08  & 0.15/0.11/0.08  & 0.16/0.12/0.07  \\
\midrule
0-6     &                           & 7               & 8               & 9               \\
\midrule
        & AUROC(95\%CI)             & 0.92(0.90-0.95) & 0.90(0.88-0.91) & 0.88(0.86-0.90) \\
        & Temporal validation       & 0.92(0.91-0.94) & 0.89(0.87-0.90) & 0.86(0.85-0.88) \\
        & Geographic validation     & 0.90(0.88-0.91) & 0.87(0.85-0.89) & 0.84(0.83-0.86) \\
        & Net Benefit(20/40/60)     & 0.16/0.11/0.08  & 0.17/0.11/0.08  & 0.16/0.09/0.07  \\
\midrule
0-7     &                           & 8               & 9               & 10              \\
        & AUROC(95\%CI)             & 0.93(0.92-0.94) & 0.91(0.88-0.92) & 0.87(0.85-0.89) \\
        & Temporal validation       & 0.92(0.90-0.93) & 0.90(0.88-0.91) & 0.85(0.82-0.87) \\
        & Geographic validation     & 0.90(0.89-0.92) & 0.88(0.86-0.90) & 0.84(0.81-0.85) \\
        & Net Benefit(20/40/60)     & 0.16/0.11/0.09  & 0.16/0.09/0.09  & 0.17/0.09/0.08  \\
\bottomrule
\end{tabular}
\caption{Predictive performance across six different observation windows. Results are shown for prediction for the next 1,2,and 3 years. The net benefits are shown at three thresholds of 20,40, and 60 percent.}
\label{tab:extended_Result}
\end{table*}

\section{TAM Questionnaire}
\label{q}

Based on a Likert scale applied to each set of questions, please respond on how strongly you agree or disagree with these statements about the early childhood obesity prediction tool.

1=Strongly disagree, 2=Disagree, 3=Neither agree nor disagree, 4=Agree, 5=Strongly Agree

\begin{enumerate}

\item  The use of the tool could help me identify patients at risk for obesity and have conversations with families about healthy lifestyles.
\item  I think it would be easy to identify patients at risk for obesity using the tool.
\item  The use of the tool could improve my identification of patients at risk for obesity and conversations with families about healthy lifestyles.
\item 	The use of the tool is compatible with my workflow.
\item 	The use of the tool could promote good clinical practice.
\item 	The use of the tool could improve my performance in identifying patients at risk for obesity and having conversations with families about healthy lifestyles.
\item 	The use of the tool could facilitate identifying patients at risk for obesity and having conversations with families about healthy lifestyles.
\item 	I think the tool would be easy to use.
\item 	I think the tool is a flexible technology to interact with.
\item 	I think I could easily learn how to use the tool.
\item 	Using the tool could help me get the most out of my time having conversations with families about healthy lifestyles.
\item 	I have the intention to use the tool for patient care.
\item 	I have the intention to use the tool when it becomes available in my clinic.
\end{enumerate}

\section{Interview Guide for User Testing of Obesity Prediction Tool}
\label{apx:int}

\subsection{Introduction}

\textbf{If e-consent is conducted right before a session, then this part can be skipped.}

Hi, thank you for taking the time to meet with me. As a reminder, during this session you will be asked to use a prototype of an electronic clinical decision support tool that will give you an estimate of a patient’s risk of obesity. The tool uses machine learning models and data from the electronic health record to predict an individual patient’s risk of developing obesity over the next three years. Because we know that early identification of obesity risk is critical but often challenging, we have designed the tool to predict obesity risk in children under the age of seven.  

During this session, we’ll show you the prototype tool and ask that you describe everything you are thinking and doing as you navigate using the tool. We are not testing you or your knowledge.  We’re testing how easy or hard it is to use the tool. At the end of the session, we will ask you a few additional questions about the tool and how useful it is. Keep in mind, this is not the final version and we are interested in getting feedback for how to improve the tool.

This session should be 30-60 minutes long and will be recorded. We will delete any personal information from our transcript of the recording and will delete the recording after the study is complete. As a reminder, this is completely voluntarily so we can stop at any point and you don’t have to answer any question that you are uncomfortable with. You will receive a \$50 dollar gift card after completing the session.
\begin{itemize}
    \item Do you still agree to be in this study?
    \item Do you have any questions for me before we start?
\end{itemize}

\subsection{Introductory Questions}

Alright, so with your permission, I will go ahead and start recording. And I am just going to say something into the recording for my records before we start with the questions. **Hit Record**. The recording has started. Today’s date is [Date] and I am here with participant [\#].

First, we’d like to know a little about your approach to discussing healthy lifestyle behaviors with families of young children.

\begin{itemize}
    \item What is your typical approach to discussing healthy lifestyle behaviors with families of young children?
    \begin{itemize}
    \item How often do you have these discussions?  What situations prompt you to have these discussions?
    \item How do you bring up the subject to families?  What communication style do you use?
    \item What topics do you cover?  What resources do you provide?
\end{itemize}
\end{itemize}
\begin{itemize}
    \item What are barriers to having these discussions?
    \begin{itemize}
    \item Do you feel like families are receptive to having these discussions?  Why or why not?
    \item Do you feel like you have enough time to have these discussions? Why or why not?
    \item Do you feel like you have the right information to have these discussions?  Why or why not?
\end{itemize}
\end{itemize}

\subsection{Prototype Tool Usability}

Thank you for sharing that with me! Now we are going to move onto reviewing the prototype tool. I am pulling it up on the screen now.  

Now, let’s envision that you are seeing patient X for a well child visit and the following information is available to you. Please walk me through what you see.

\textbf{Explanation of tool}
\begin{itemize}
    \item What do you think about the statement at the top?
        \item How do you interpret it?
    \item Is it easy to understand?  Why or why not?  
    \item Is it helpful?  Why or why not?
    \item Is there anything you would change about the statement?
\end{itemize}

\textbf{Obesity risk graphs}
\begin{itemize}
    \item What do you think about the graphs?
    \item How do you interpret them?
    \item Is it easy to understand?  Why or why not?
    \item Is it helpful?  Why or why not?
    \item Is there anything you would change about the graphs?
\end{itemize}

\textbf{Risk factors}
\begin{itemize}
    \item What do you think about the description of risk factors?
    \item How do you interpret it?
    \item Is it easy to understand?  Why or why not?  
    \item Is it helpful?  Why or why not?
    \item Is there anything you would change about the table?
\end{itemize}

\textbf{Healthy lifestyle resources}
\begin{itemize}
    \item What do you think about the resources provided?
    \item Are they helpful?  Why or why not?
    \item Is there anything you would change about the resources section?
\end{itemize}

\subsection{Prototype Tool Acceptability and Helpfulness}
Great! Thank you for going through the tool. Before we finish today, I just want to ask you a few extra questions.

\begin{itemize}
    \item What is your overall impression of the tool?
    \begin{itemize}
    \item What did you like best/least about the tool?
\end{itemize}
\item How useful would you find the tool?
\begin{itemize}
    \item How much do you think the tool would inform your discussions with families about healthy lifestyles?
    \item Do you think the tool would help you feel more confident knowing what families might need additional counseling and resources related to healthy lifestyles?  Why or why not?
    \item Do you think the tool would help you feel more comfortable bringing up the subject to families?  Why or why not?
\end{itemize}
\item How would you see yourself using the tool?  
\begin{itemize}
    \item How often would you use it?
    \item What would make you more/less likely to use the tool?
    \item Where in your workflow would it be most helpful to have the tool?
\end{itemize}

\item Is there anything you would change/add to the tool?
\begin{itemize}
    \item For example, would it be helpful to have a growth curve showing a child’s future predicted BMI trajectory?
    \item What barriers would you have or think other providers might have to using the tool?
\end{itemize}
\end{itemize}

Finally, do you have any last thoughts or feedback to give to the team to improve the tool?

\subsection{Closing}

Thank you for sharing this with me!   That’s all I have for today. I will turn off the recording now. 

I am just going to confirm your mailing address so I know where to send the gift card.  As a reminder, at Nemours, we use a gift card company called Greenphire. We will share your name, date of birth, and mailing address with them so they can load your gift card with money.

\end{document}